\def\publicpreprint{1}
\def\SupplementaryVideoURL{https://youtu.be/-JTRXvstFAo}
\def\SupplementaryVideoLabel{youtu.be/-JTRXvstFAo}

\documentclass[letterpaper, 10 pt, conference]{ieeeconf}  %

\IEEEoverridecommandlockouts                              %
\usepackage{graphicx} %
\usepackage{epsfig} %
\usepackage{mathptmx} %
\usepackage{times} %
\usepackage{amsmath} %
\usepackage{amssymb}  %
\usepackage{mathtools} %
\usepackage{xcolor,colortbl}
\usepackage{booktabs}
\usepackage{tabularx}
\usepackage{multirow}
\usepackage{algorithm}
\usepackage{algpseudocode}
\usepackage{url}
\usepackage{cite}
\usepackage{setspace}
\usepackage{tikz}
\usepackage{bm}
\usepackage[hidelinks]{hyperref}
\usepackage{xurl}
\usetikzlibrary{arrows.meta, positioning, calc, decorations.pathreplacing, fit, backgrounds}

\usepackage[caption=false,font=footnotesize]{subfig}
\usepackage{stfloats} %

\newcommand{\tabnote}[1]{\par\vspace{5pt}\noindent\parbox{\columnwidth}{\scriptsize #1}}
\DeclareRobustCommand{\stock}{Stock}
\DeclareRobustCommand{\high}{High}
\DeclareRobustCommand{\highv}{\mbox{High-V}}

\makeatletter
\let\oldthebibliography\thebibliography
\renewcommand{\thebibliography}[1]{%
  \oldthebibliography{#1}%
  \setlength{\itemsep}{0pt}%
  \setlength{\parskip}{0pt}%
}
\makeatother

\title{\LARGE \bf
Battery-Aware Reinforcement Learning for Aggressive Quadrotor Flight
}

\ifdefined\publicpreprint
\author{Alejandro S\'anchez Roncero, Olov Andersson and Petter \"Ogren%
\thanks{This work was partially supported by the Wallenberg AI, Autonomous
Systems and Software Program (WASP) funded by the Knut and Alice Wallenberg
Foundation. The authors are with the Department of Robotics, Perception and Learning,
School of Electrical Engineering and Computer Science, Royal Institute of
Technology (KTH), SE-100 44 Stockholm, Sweden. Corresponding author:
\href{mailto:alesr@kth.se}{\texttt{alesr@kth.se}}.}}
\hypersetup{pdftitle={Battery-Aware Reinforcement Learning for Aggressive Quadrotor Flight},
pdfauthor={Alejandro S\'anchez Roncero, Olov Andersson, Petter \"Ogren}}

\let\publicbibliography\thebibliography
\renewcommand{\thebibliography}[1]{\publicbibliography{#1}\small\setlength{\itemsep}{2pt}}
\IEEEtriggeratref{17}

\else
\author{Anonymous Authors}
\fi

\definecolor{lavender}{rgb}{0.9, 0.9, 0.98}
\definecolor{red}{RGB}{255, 0, 0}
\definecolor{orange}{RGB}{252, 130, 62}
\definecolor{blue}{RGB}{0, 0,255}
\definecolor{darkgreen}{RGB}{0, 150,0}

\newcommand{\todo}[1]{\remark{TODO}{red}{#1}}

\newcommand{\remark}[3]{{\color{#2}[#1: #3]}}

\ifdefined\paperreading
  \renewcommand{\todo}[1]{}
\fi

\begin{document}

\maketitle
\thispagestyle{empty}
\pagestyle{empty}

\suppressfloats[t]
\begin{abstract}
Agile flight tasks such as drone racing and pursuit--evasion require strong acceleration and precise turns, but the available thrust changes as the battery discharges and voltage drops under load. Conservative command limits make this variation easier to tolerate, at the cost of unused performance. We investigate how learned controllers can use that additional thrust while retaining the flight controller's voltage compensation and rate control. Our training simulator couples an identified load-transient battery model to rotor dynamics and firmware saturation. The feedforward policy receives filtered voltage during both training and deployment. Controlled ablations distinguish the benefit of a larger thrust-command range from that of voltage information. On a 38~g Crazyflie Brushless, the resulting policy reduces circle tracking error by 49\% relative to stock-authority RL at 3.84~m/s, while preserving easy-task precision. Mean 20-lap race time decreases from 106.22~s to 95.24~s. Compared with a voltage-blind policy with the same increased authority, hardware error and race time are lower by 15.3\% and 4.5\%, respectively. In simulation, replacing the policy's voltage input with a recording from a different battery condition worsens hard-circle tracking, with a smaller, voltage-dependent effect in racing. Together, these results show where a simple voltage input complements existing actuator compensation in aggressive learned flight.
\end{abstract}
\begin{keywords}
Aerial Systems: Mechanics and Control, Reinforcement Learning, Calibration and Identification
\end{keywords}

\ifdefined\publicpreprint
  \ifx\SupplementaryVideoURL\empty\else
    \par\smallskip\noindent\textit{Supplementary video: }
    \href{\SupplementaryVideoURL}{\expandafter\nolinkurl\expandafter{\SupplementaryVideoLabel}}\par
  \fi
\fi

\section{INTRODUCTION}\label{sec:introduction}
Quadrotors trained with reinforcement learning (RL) can race at champion level~\cite{kaufmann2023champion} and learn their path and speed directly from a racing objective~\cite{songReachingLimitAutonomous2023}. Related approaches enable agile pursuit--evasion and competitive maneuvers~\cite{roncero2025learned,pasumarti2025agile}. Their performance depends on available thrust, which falls as the battery discharges and drops further under high load. A maneuver that is feasible early in a flight can therefore become harder to execute later.

\begin{figure}[t]
\centering
\includegraphics[width=\columnwidth]{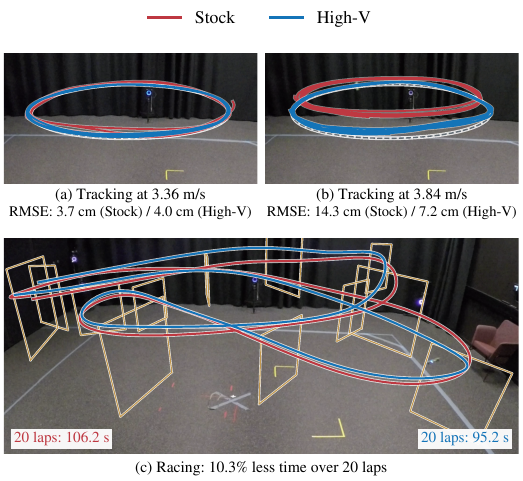}
\caption{Hardware flights with \stock{} (default thrust-command limit, no voltage input) and \highv{} (higher limit with voltage input). (a,b) Tracking at $3.36$ and $3.84$\,m/s, with root-mean-square error (RMSE) averaged over repeated flights. (c) Racing over $20$ laps. The overlays show measured paths on calibrated camera images and virtual gates in racing.}
\label{fig:first_figure}
\end{figure}

Flight controllers partly address this variation through voltage compensation. Betaflight's sag compensation and ArduPilot's voltage scaling are related examples~\cite{betaflightSagCompensation,ardupilotVoltageScaling}. Our controller raises motor duty as voltage falls to keep thrust response approximately consistent. Conservative command limits leave room for attitude corrections, but can keep useful thrust out of reach. Raising the limit makes saturation more important to the policy.

Thrust and attitude control share the same four motors. A sharp turn may saturate one motor even when the requested total thrust seems feasible, forcing the mixer to reduce or redistribute commands~\cite{faessler2016thrust}. Battery sag further lowers what these commands deliver, so the policy must account for both demands on the motors.

Exploiting the available thrust requires an electromechanical view of the quadrotor, connecting battery supply to motor response and the resulting forces~\cite{gupta2026bc}. We bring this view into RL training by modeling how motor load changes voltage, how voltage changes rotor speed, and how the firmware compensates and saturates. A feedforward policy uses a relaxed thrust-command limit and one filtered voltage input during both training and flight, while still commanding collective thrust and rates through the existing onboard controller. Alongside motion feedback, which reflects recent actions, voltage can help identify changes in available thrust before large tracking errors develop.

All baselines also train with the battery model, separating the effect of observing voltage from that of simulating it. We ask: \emph{does additional command authority improve flight, and does observing voltage help a policy use that authority?} \stock{} uses the default thrust-command limit and \high{} raises it, both without voltage input. \highv{} adds a voltage input with a $10$\,s filter to \high{}. We test tracking accuracy on fixed-speed circles and evaluate the policies' own choices of path and speed in racing (Fig.~\ref{fig:first_figure}).

Our contributions are:
\begin{itemize}
\clubpenalty=10000
\item A battery-conditioned control design that combines increased thrust-command authority with filtered voltage input, while retaining onboard compensation and rate control. Separate comparisons establish the benefits of additional authority and voltage information.
\item A battery-aware actuation model for RL training that captures how discharge and rapid load changes affect available thrust, including onboard compensation and motor saturation. We identify the model from measurements and validate its predictions against flight data.
\item Hardware validation on the Crazyflie Brushless in circle tracking and drone racing, demonstrating improved performance over both stock-authority and voltage-blind high-authority policies.
\end{itemize}

\section{RELATED WORK}\label{sec:related}
\paragraph{Battery-dependent modeling and control}
Battery-dependent modeling links electrical behavior to the forces available for flight. Thrust dynamics can be identified from motor commands and measured terminal voltage~\cite{staub2015battery}, but predicting future performance also requires accounting for voltage changes under load. Equivalent-circuit models capture these transients~\cite{chen2006accurate}, and coupled battery, motor, and aerodynamic models extend prediction to range and endurance~\cite{bauersfeldRangeEnduranceOptimal2022}. These effects also matter for control, whether balancing altitude-tracking performance against energy use~\cite{tavares2025tradeoff}, compensating voltage-induced force errors~\cite{liu2025accurate}, or anticipating propulsion limits with battery-constrained nonlinear model predictive control (NMPC)~\cite{gupta2026bc}. We focus on rapid load changes in aggressive flight, predicting voltage from duty and rotor-speed history without a current sensor. Our added load-onset term improves high-load prediction on held-out flights.

\paragraph{Actuation in learned flight}
In learned racing, the policy can choose its path and speed by optimizing gate progress rather than following a timed reference~\cite{songReachingLimitAutonomous2023}. The simulated actuator response therefore shapes the maneuvers it learns. Swift already represents battery effects by coupling a power-driven battery model and voltage-dependent rotor speed to Betaflight control and motor saturation~\cite{kaufmann2023champion}. That captures battery effects during training, while its reported actor inputs are vehicle state, gates, and previous action.

We investigate the value of an explicit voltage input using the identified Crazyflie Brushless model, whose motor lag and measured thrust and torque curves have supported transfer of learned controllers~\cite{grafe2026model}. We retain its rotor fits and extend the motor response to depend on voltage, alongside calibrated drag and deployed firmware compensation and desaturation. Dynamics randomization~\cite{peng2018sim} covers uncertainty around this model. This supports separate comparisons of increased thrust-command authority and voltage input with onboard compensation retained.

\paragraph{Adaptation and training information}
A controller can also respond to changing dynamics through feedback rather than an explicit voltage input. DATT supplies a learned tracker with a deployed $\mathcal L_1$ disturbance estimate~\cite{huang2023datt}, while adaptation modules can infer dynamics from motion history~\cite{kumar2021rma}. In racing, SkyDreamer uses a recurrent world model to estimate attainable rotor speed and adjust its path as the battery depletes, commanding motors directly~\cite{verraest2025skydreamer}. We study a scalar voltage input to a model-free feedforward actor that retains the existing collective-thrust/rate controller.

Information used during flight is distinct from information that only aids learning. Asymmetric actor--critic training gives simulator information to a critic discarded before deployment~\cite{pinto2017asymmetric}. We therefore evaluate critic privilege separately from the voltage input available to the actor in flight.

\section{METHOD}\label{sec:method}
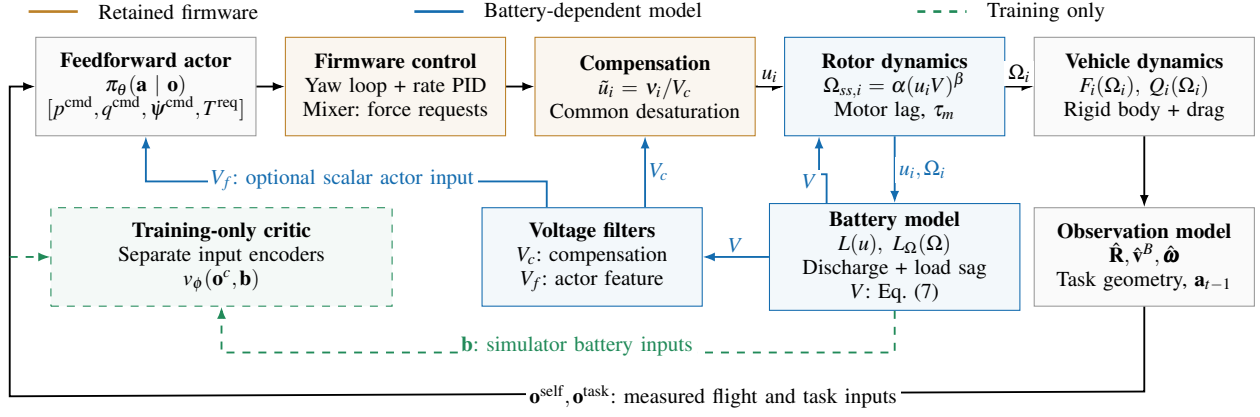
\begin{figure*}[t]
\vspace*{2pt} %
\centering
\definecolor{modelblue}{HTML}{146CAF}
\definecolor{inversegold}{HTML}{BC8128}
\definecolor{criticgreen}{HTML}{218A59}
\begin{tikzpicture}[x=1cm,y=1cm,font=\footnotesize,
 box/.style={draw=black!40,fill=black!2,align=center,text width=27mm,minimum height=13mm,inner sep=3pt},
 model/.style={box,draw=modelblue,fill=modelblue!5},
 firmware/.style={box,draw=inversegold,fill=inversegold!7},
 arrow/.style={-{Latex[length=1.6mm]},line width=.7pt},
 sig/.style={font=\footnotesize,fill=white,inner sep=1.3pt}]
\node[box] (actor) at (0,0) {\textbf{Feedforward actor}\\$\pi_\theta(\mathbf a\mid\mathbf o)$\\$[p^{\rm cmd},q^{\rm cmd},\dot\psi^{\rm cmd},T^{\rm req}]$};
\node[firmware] (pid) at (3.3,0) {\textbf{Firmware control}\\Yaw loop + rate PID\\Mixer: force requests};
\node[firmware] (inverse) at (6.6,0) {\textbf{Compensation}\\$\tilde u_i=\nu_i/V_c$\\Common desaturation};
\node[model] (motor) at (9.9,0) {\textbf{Rotor dynamics}\\$\Omega_{ss,i}=\alpha(u_iV)^\beta$\\Motor lag, $\tau_m$};
\node[box] (plant) at (13.2,0) {\textbf{Vehicle dynamics}\\$F_i(\Omega_i),\;Q_i(\Omega_i)$\\Rigid body + drag};
\draw[arrow] (actor) -- (pid);
\draw[arrow] (pid) -- (inverse);
\draw[arrow] (inverse) -- node[sig,above]{$u_i$} (motor);
\draw[arrow] (motor) -- node[sig,above]{$\Omega_i$} (plant);
\node[model,text width=31mm] (battery) at (9.9,-2.25) {\textbf{Battery model}\\$L(u),\;L_\Omega(\Omega)$\\Discharge + load sag\\$V$: Eq.~\eqref{eq:battery}};
\node[box] (obs) at (13.2,-2.25) {\textbf{Observation model}\\$\hat{\mathbf R},\hat{\mathbf v}^B,\hat{\boldsymbol\omega}$\\Task geometry, $\mathbf a_{t-1}$};
\node[model,text width=27mm] (filter) at (5.9,-2.25) {\textbf{Voltage filters}\\$V_c$: compensation\\$V_f$: actor feature};
\node[box,draw=criticgreen,fill=criticgreen!5,dashed,text width=43mm] (critic) at (1.0,-2.25)
 {\textbf{Training-only critic}\\Separate input encoders\\$v_\phi(\mathbf o^c,\mathbf b)$};
\draw[arrow] (plant) -- (obs);
\draw[arrow,modelblue] (motor.south) -- node[sig,right]{$u_i,\Omega_i$} (battery.north);
\draw[arrow,modelblue] ([xshift=-9mm]battery.north) -- ++(0,.38) -| ([xshift=-10mm]motor.south);
\node[sig,text=modelblue] at (8.78,-1.25) {$V$};
\draw[arrow,modelblue] (battery.west) -- node[sig,above]{$V$} (filter.east);
\draw[arrow,modelblue] ([xshift=7mm]filter.north) -- node[sig,right]{$V_c$} (inverse.south);
\draw[arrow,modelblue] ([xshift=-6mm]filter.north) -- ++(0,.34) -| (actor.south);
\node[sig,text=modelblue] at (2.6,-1.23) {$V_f$: optional scalar actor input};
\draw[arrow,dashed,criticgreen] (battery.south) -- ++(0,-.55) -| (critic.south);
\node[sig,text=criticgreen] at (5.7,-3.43) {$\mathbf b$: simulator battery inputs};
\draw[arrow] (obs.south) -- ++(0,-1.12) -- (-1.80,-4.09) -- (-1.80,0) -- (actor.west);
\node[sig] at (7.5,-4.09) {$\mathbf o^{\rm self},\mathbf o^{\rm task}$: measured flight and task inputs};
\draw[arrow,dashed,criticgreen] (-1.80,-2.25) -- (critic.west);
\draw[inversegold,line width=1pt] (-1.55,.99) -- (-.90,.99);
\node[anchor=west,font=\footnotesize] at (-.75,.99) {Retained firmware};
\draw[modelblue,line width=1pt] (3.55,.99) -- (4.20,.99);
\node[anchor=west,font=\footnotesize] at (4.35,.99) {Battery-dependent model};
\draw[criticgreen,dashed,line width=1pt] (10.2,.99) -- (10.85,.99);
\node[anchor=west,font=\footnotesize] at (11.0,.99) {Training only};
\end{tikzpicture}
\setlength{\abovecaptionskip}{0pt} %
\caption{The simulator reproduces the deployed control path. Motor load changes voltage, affecting rotor speed and firmware compensation ($V_c$). The actor receives filtered voltage ($V_f$), while the training-only critic also receives simulator battery inputs $\mathbf b$ (Section~\ref{sec:battery_modeling}).}
\label{fig:pipeline}
\end{figure*}

Our simulator links motor load, battery voltage, and available thrust (Fig.~\ref{fig:pipeline}). Numerical settings appear in Section~\ref{sec:setup}.

\subsection{Vehicle and Rotor Dynamics}
Let $\mathbf p,\mathbf v$ denote world position and velocity, $\mathbf R\in SO(3)$ the body-to-world rotation, $\boldsymbol\omega$ body rates, and $\Omega_i$ the speed of rotor $i\in\{1,\ldots,4\}$. With mass $m$, body inertia $\mathbf J$, and gravity $\mathbf g=[0,0,-g]^\top$,
\begin{align}
\dot{\mathbf p}&=\mathbf v,&
m\dot{\mathbf v}&=m\mathbf g+\mathbf R(\mathbf f_{\rm prop}+\mathbf f_{\rm aero}),\nonumber\\
\dot{\mathbf R}&=\mathbf R[\boldsymbol\omega]_\times,&
\mathbf J\dot{\boldsymbol\omega}&=\boldsymbol\tau_{\rm prop}-\boldsymbol\omega\times\mathbf J\boldsymbol\omega.
\label{eq:dynamics}
\end{align}
Here $g$ is gravitational acceleration, $[\cdot]_\times$ the cross-product matrix, and $\mathbf e_z=[0,0,1]^\top$ the body thrust axis. Thrust is $\mathbf f_{\rm prop}=\mathbf e_z\sum_i F_i(\Omega_i)$, using measured cubic fits for rotor thrust $F_i$ and aerodynamic/friction torque $Q_i$~\cite{grafe2026model}. With body-frame rotor position $\mathbf r_i$, rotor inertia $J_r$, and reaction-torque sign $s_i\in\{-1,1\}$, total rotor moment is
\begin{equation}
\boldsymbol\tau_{\rm prop}=\sum_i\left[\mathbf r_i\times(F_i\mathbf e_z)+s_i(Q_i+J_r\dot\Omega_i)\mathbf e_z\right].
\label{eq:rotor_moment}
\end{equation}
Body-frame drag is
$\mathbf f_{\rm aero}=-(\sum_i\Omega_i)\mathbf K_{\rm aero}\mathbf R^\top\mathbf v$, with diagonal drag coefficients $\mathbf K_{\rm aero}$.
The Crazyflie Brushless model~\cite{grafe2026model} uses $\dot\Omega_i=(Ku_i-\Omega_i)/\tau_m$, with motor duty $u_i\in[0,1]$, gain $K$, and time constant $\tau_m$. We retain this lag and the thrust and torque curves, but replace $Ku_i$ with a voltage-dependent target (Section~\ref{sec:actuation}).

\subsection{Voltage-Conditioned Actuation}\label{sec:actuation}
We use terminal voltage $V$, which includes the drop under motor load. Fitted gain $\alpha$ and exponent $\beta$ set steady-state rotor speed $\Omega_{ss,i}$:
\begin{align}
\Omega_{ss,i}&=\alpha(u_iV)^\beta,\nonumber\\
\Omega_{i,k+1}&=\Omega_{ss,i}+(\Omega_{i,k}-\Omega_{ss,i})e^{-\Delta t/\tau_m}.
\label{eq:voltage_rotor}
\end{align}
Here $k$ indexes steps of duration $\Delta t$. Even at full duty, rotor speed now depends on voltage.

\paragraph{Command interface}
The clipped action $\mathbf a\in[-1,1]^4$ is decoded as
\begin{equation}
\begin{split}
[p^{\rm cmd},q^{\rm cmd},\dot\psi^{\rm cmd}]&=\mathbf s_\omega\odot\mathbf a_{1:3},\\
T^{\rm req}&=\tfrac12(a_4+1)T_{\rm scale}.
\end{split}\label{eq:action}
\end{equation}
Here $\odot$ is elementwise multiplication, $\mathbf s_\omega$ gives rate scales, and $T_{\rm scale}$ scales requested collective thrust $T^{\rm req}$. Commands $p^{\rm cmd},q^{\rm cmd}$ target roll/pitch rates. Firmware integrates yaw velocity $\dot\psi^{\rm cmd}$ into heading for its outer yaw controller. We retain command quantization, clipping, and the deployment yaw-sign conversion.

Bitcraze's firmware converts mixed motor counts to force requests using a per-motor scale~\cite{bitcrazeMotorCompensation}. Raising it from $0.20$ to $0.25$\,N unlocks more of the existing hardware capability, with the corresponding actor scales in Table~\ref{tab:contracts}. The physical motor model is unchanged, but conversion scales both collective and PID differential requests. \high{} and \highv{} share this interface.

\paragraph{Compensation and saturation}
Rate proportional--integral--derivative (PID) control and motor mixing turn these commands into per-motor force requests. Inverting a nominal cubic thrust curve gives required motor voltage $\nu_i$. Compensation converts this to duty $\tilde u_i=\nu_i/V_c$, using filtered supply $V_{c,k}=a_cV_{c,k-1}+(1-a_c)V_k$ with filter weight $a_c$. If a request exceeds duty limit $u_{\max}$, common desaturation removes the largest excess $\delta$ from every motor:
\begin{equation}
\delta=\max(0,\max_i\tilde u_i-u_{\max}),\qquad
u_i=\operatorname{clip}(\tilde u_i-\delta,0,u_{\max}).
\label{eq:desaturation}
\end{equation}
This sacrifices collective thrust to accommodate differential commands, but does not exactly preserve physical torque because force is nonlinear in duty. The modeled PID retains derivative on measured rate, integer truncation, output clipping, and integral clamping, without mixer-aware back-calculation. Importantly, compensation uses filtered $V_c$, while the motors respond to terminal $V$. Physical motor parameters are randomized during training, but the firmware inverse stays nominal, as on the vehicle.

\begin{figure}[b]
\centering
\includegraphics[width=\columnwidth]{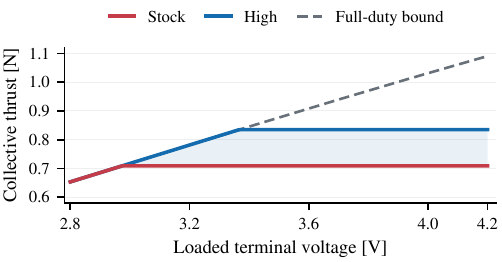}
\caption{Modeled thrust limits with equal motor commands and settled compensation ($V_c=V$). Solid curves use the maximum host command ($60000$ counts), with force scales of $0.8$~N for \stock{} and $1.0$~N for \high{}. The dashed curve assumes all four motors at full duty. Attitude corrections can reduce the available collective thrust.}
\label{fig:thrust_limits}
\end{figure}

Fig.~\ref{fig:thrust_limits} shows the resulting thrust limits under settled compensation.

\subsection{Load-Transient Battery Model}\label{sec:battery_modeling}
Battery voltage falls gradually as the pack is used, but also drops temporarily when motor load rises. We model both effects from duty and rotor speed without measuring current. The internal states summarize load history rather than electrochemical state of charge.

We estimate load with $L=\sum_i u_i^\rho$ and $L_\Omega=\sum_i(\Omega_i/\bar\Omega)^2$, where $\rho$ is a fitted exponent and $\bar\Omega$ a reference rotor speed. Accumulated load $q_k=q_{k-1}+\Delta tL_k$ captures depletion. To capture sag and recovery, states $z_j$, $j\in\{d,r,z,h\}$, filter inputs $x_j$ with time constants $\tau_j$:
\begin{equation}
z_{j,k}=a_jz_{j,k-1}+(1-a_j)x_{j,k},\quad a_j=e^{-\Delta t/\tau_j}.
\label{eq:filter}
\end{equation}
The duty load drives both $z_d$ and $z_z$, with $x_d=x_z=L$. Different time constants, $\tau_d<\tau_z$, let them capture fast and slow voltage responses. The other inputs are rotor load $x_r=L_\Omega$ and squared duty load $x_h=L^2$. After updating these states, terminal voltage is
\begin{equation}
\begin{split}
V_k=\operatorname{clip}\bigl(&V_0-k_q q_k-k_dz_{d,k}-k_rz_{r,k}-k_zz_{z,k}\\
&-k_h\max(L_k^2-z_{h,k},0),\ V_{\min},V_{\max}\bigr).
\end{split}\label{eq:battery}
\end{equation}
The fitted coefficients $k_q,k_d,k_r,k_z,k_h$ scale voltage drops from reference $V_0$, with limits $V_{\min},V_{\max}$. The last term adds extra sag while load exceeds its recent history and vanishes at steady load. It captures sharp voltage dips without forcing a recovery overshoot (Section~\ref{sec:identification}).

At reset, $q=(V_0-V_{\rm reset})/k_q$ and $z_j=0$, giving rested-equivalent voltage $V_{\rm reset}$, not an in-flight open-circuit measurement. The simulator battery vector $\mathbf b=(V,V_0,q,z_d,z_r,z_z,L)\in\mathbb R^7$ is an optional critic input. Only filtered voltage is available to the actor when enabled (Section~\ref{sec:learning}).

\subsection{Policy and Training Information}\label{sec:learning}
Each task is a partially observable Markov decision process (POMDP)~\cite{kaelbling1998planning}. At step $t$, state $\mathbf s_t$ includes vehicle, task, motor and battery variables, controller and observer histories, delays, and randomized parameters. Dynamics give $\mathbf s_{t+1}\sim P(\cdot\mid\mathbf s_t,\mathbf a_t)$, while observations $\mathbf o_t\sim Z(\cdot\mid\mathbf s_t)$ expose flight measurements, task geometry, and optional voltage. Battery state affects available force even when unobserved.

A Gaussian feedforward actor $\pi_\theta(\mathbf a_t\mid\mathbf o_t)$ maps the current observation to an action. We train it with proximal policy optimization (PPO)~\cite{schulman2017ppo} to maximize expected discounted reward:
\begin{equation}
\max_\theta J(\theta),\qquad
J(\theta)=\mathbb E_{\pi_\theta}\!\left[\sum_{t=0}^{\infty}\gamma^t r_t\right].
\label{eq:actor_objective}
\end{equation}
Here $r_t$ is reward, $\gamma\in[0,1)$ the discount, and the expectation covers policy-generated training episodes. Training samples actions, whereas deployment uses their clipped mean. Timeouts bootstrap the value estimate, but failures terminate it.

Both tasks share measured flight inputs
\begin{equation}
\mathbf o_t^{\rm self}=[\operatorname{vec}(\hat{\mathbf R}_t),\hat{\mathbf v}_t^B,
\hat{\boldsymbol\omega}_t,\mathbf a_{t-1}]\in\mathbb R^{19}.
\label{eq:shared_obs}
\end{equation}
Hats denote estimates, $B$ body coordinates, $\operatorname{vec}$ matrix flattening, and $\mathbf a_{t-1}$ the previous action. Circle task inputs are
\begin{equation}
\mathbf o_t^{\rm track}=[\mathbf e_{p,t}^B,\Delta\mathbf p_{1:6,t}^B,
\cos\tilde\psi_t,\sin\tilde\psi_t]\in\mathbb R^{23},
\label{eq:track_obs}
\end{equation}
where $\mathbf e_{p,t}^B=\hat{\mathbf R}_t^\top(\mathbf p_{\rm ref}(t)-\hat{\mathbf p}_t)$,
$\Delta\mathbf p_{j,t}^B=\hat{\mathbf R}_t^\top[\mathbf p_{\rm ref}(t+h_j)-\mathbf p_{\rm ref}(t)]$, with reference position $\mathbf p_{\rm ref}$, preview offsets $h_j$, and desired-minus-measured yaw $\tilde\psi_t$. Thus previews describe future reference-position increments, not reference velocity or acceleration. Racing instead observes the four corners of each of the next two gates:
\begin{equation}
\mathbf o_t^{\rm race}=
\bigl[\hat{\mathbf R}_t^\top(\mathbf c_{g,j}-\hat{\mathbf p}_t)\bigr]_{g=g_t,g_t+1;\,j=1:4}
\in\mathbb R^{24}.
\label{eq:race_obs}
\end{equation}
Here $g_t$ is the active gate and $\mathbf c_{g,j}$ its world-frame corner. Gate-relative inputs and progress objectives follow established racing designs~\cite{kaufmann2023champion,songReachingLimitAutonomous2023}.

The actor's observation $\mathbf o$ combines flight inputs, $\mathbf o^{\rm task}$ (tracking or racing), and optional filtered voltage $V_f$, simulated during training and measured in flight. We compare $54$\,ms filtering with slower variants, using $10$\,s in \highv{}. Filters initialize from startup voltage. Previous action starts at hover thrust and zero rates, and input normalization is frozen after training. A constant-voltage control tests whether the extra input dimension alone explains a gain.

The training-only critic estimates discounted future reward. A symmetric critic receives the actor's observations, while a battery-privileged critic also uses $\mathbf b_t$ and learns parameters $\phi$ to approximate
\begin{equation}
v_\phi(\mathbf o_t^c,\mathbf b_t)\approx
\mathbb E_{\pi_\theta}\!\left[\sum_{\ell=0}^{\infty}\gamma^\ell r_{t+\ell}\,\middle|\,\mathbf o_t^c,\mathbf b_t\right].
\label{eq:critic_value}
\end{equation}
The critic input $\mathbf o_t^c$ shares the actor's flight and task observations. We process $\mathbf o_t^c$ and $\mathbf b_t$ with separate networks, then combine their outputs to predict the scalar value without requiring the full hidden state.

\subsection{Tasks and Rewards}\label{sec:tasks}
\paragraph{Circle tracking}
A counter-clockwise circle with fixed heading tests tracking at a prescribed speed. For radius $r$ and speed $v$, balancing gravity and centripetal acceleration requires ideal collective $T_{\rm ideal}=m\sqrt{g^2+(v^2/r)^2}$, before drag and differential-control losses. We train a separate policy for each speed. Its reward encourages accurate position and heading tracking while penalizing large body rates and abrupt action changes:
\begin{equation}
\begin{split}
r_t^{\rm track}={}&w_p[(1-\lambda_p)e^{-k_1e_t^2}+\lambda_pe^{-k_2e_t^2}]
-w_\psi(1-\cos e_{\psi,t})\\
&-w_\omega(\|\boldsymbol\omega_{1:2,t}\|_2+|\omega_{z,t}|)
-w_a(\|\Delta\mathbf a_{1:3,t}\|_2+|\Delta a_{4,t}|).
\end{split}\label{eq:track_reward}
\end{equation}
Here $e_t=\|\mathbf p_t-\mathbf p_{\rm ref}(t)\|$ is position error and $e_{\psi,t}$ is true heading error. The two position terms reward tracking at different error scales, set by $k_1,k_2$ and mixed by $\lambda_p$. The remaining terms use roll/pitch rates $\boldsymbol\omega_{1:2,t}$, yaw rate $\omega_{z,t}$, and raw clipped action changes $\Delta\mathbf a_t=\mathbf a_t-\mathbf a_{t-1}$. The $w$ coefficients weight these contributions. Excessive tracking error, invalid state, or arena exit ends the episode with a failure cost.

\paragraph{Gate racing}
Racing lets the policy choose its path and speed, provided it crosses the gates through their apertures in the prescribed direction and order. The reward combines progress toward the next gate and a valid-crossing bonus, while penalizing large body rates and deviations from the course:
\begin{equation}
\begin{split}
r_t^{\rm race}={}&w_d[d_{g_t}(\mathbf p_{t-1})-d_{g_t}(\mathbf p_t)]
+w_g I_{g,t}-w_r\|\boldsymbol\omega_t\|_2\\
&-w_c\left[\operatorname{clip}\left(\frac{c_t-c_0}{c_{\max}-c_0},0,1\right)\right]^2 .
\end{split}\label{eq:race_reward}
\end{equation}
Here $d_g(\mathbf p)$ is distance to gate center $g$, and $I_{g,t}$ indicates a valid crossing. Progress is measured against the same gate before its index advances. The corridor penalty uses distance $c_t$ from the path centerline, starting at $c_0$ and reaching full weight at failure distance $c_{\max}$. The four $w$ coefficients balance these objectives. Missed gates, corridor violation, contact, invalid state, and arena exit incur a failure cost. Training perturbs starts and gate centers and accepts crossings through a smaller aperture while observing full-size corners.

\section{EXPERIMENTS}\label{sec:experiments}
We first validate the battery model, then compare hardware performance and examine the effects of actor voltage input and training-only battery information.

\subsection{Platform and Learning Protocol}\label{sec:setup}
The guardless Crazyflie Brushless weighs $38$\,g including its SD logging deck and card. Motion capture provides pose at $100$\,Hz. Onboard logs record voltage, pulse-width modulation (PWM), rotor speed in revolutions per minute (RPM), and controller states at up to $500$\,Hz during identification and $250$\,Hz in final flights.

For training, Isaac Lab on Isaac Sim~\cite{mittal2025isaac} runs vehicle dynamics and firmware control at $500$\,Hz, with a $50$\,Hz actor and $10$\,ms action latency. We use SKRL's PPO implementation~\cite{serrano2023skrl} with separate actor and critic observation normalizers and the settings in Tables~\ref{tab:contracts} and~\ref{tab:plant}. All methods, including \stock{}, train with three seeds per setting.

\begin{table}[tb]
\vspace*{6pt} %
\centering\footnotesize
\caption{Learning and command settings}
\setlength{\tabcolsep}{3.5pt}
\begin{tabular}{@{}lcc@{}}
\toprule
Parameter & Circle & Racing\\
\midrule
Actor hidden layers, ELU & $256,256,128$ & $256,256,128$\\
Symmetric critic hidden layers & $256,256,128$ & $256,256,128$\\
Privileged critic: observation & $256,128$ & $256,128$\\
Privileged critic: battery & $64,64$ & $64,64$\\
Privileged critic: value & $256,128,1$ & $256,128,1$\\
\midrule
No voltage / scalar voltage & $42/43$ inputs & $43/44$ inputs\\
Training frames & $80$M & $120$M\\
Parallel environments & $6144$ & $6144$\\
Episode duration & $10$ s & $20$ s\\
Rollout / epochs / minibatches & $16/8/16$ & $32/8/8$\\
Initial learning rate & $2\!\times\!10^{-4}$ & $10^{-4}$\\
Maximum adaptive rate & $10^{-2}$ & $4\!\times\!10^{-4}$\\
KL threshold & $0.008$ & $0.008$\\
Initial log standard deviation & $-0.5$ & $-0.5$\\
Discount $\gamma$ / GAE $\lambda$ & $.995/.95$ & $.9995/.995$\\
Entropy coefficient & $0$ & $.002$\\
Policy / value clipping & $0.2/0.2$ & $0.2/0.2$\\
Gradient clip / value scale & $0.5/1.0$ & $0.5/1.0$\\
$\mathbf s_\omega$ [deg/s] & $175,175,200$ & $225,225,200$\\
\stock{} / \high{} $T_{\rm scale}$ [N] & $0.8/1.0$ & $0.8/1.0$\\
Firmware force scale, Stock/High [N] & $0.20/0.25$ & $0.20/0.25$\\
Host command cap [counts] & $60000$ & $60000$\\
Applied duty rail $u_{\max}$ & $1$ & $1$\\
Training reset voltage [V] & $3.45$--$4.20$ & $3.45$--$4.20$\\
\bottomrule
\end{tabular}
\tabnote{Primary comparisons use the privileged critic. Settings are fixed within each task, and \highv{} uses the same command scales as \high{}. ELU: exponential linear unit; KL: Kullback--Leibler divergence; GAE: generalized advantage estimation. Thrust sensitivity is one. The host cap limits requested thrust to $0.732$~N for \stock{} and $0.916$~N for \high{}, although actual thrust depends on battery and motor loading.}
\label{tab:contracts}
\end{table}

\begin{figure}[t]
\centering
\includegraphics[width=\columnwidth]{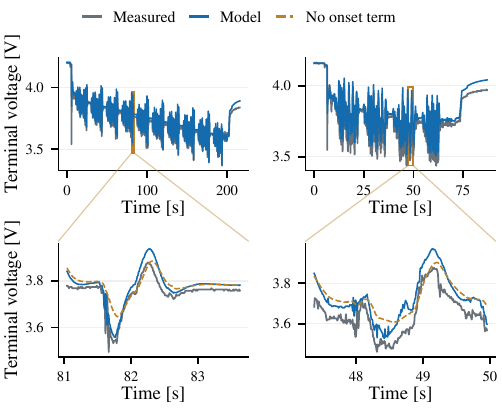}
\caption{Measured and predicted battery voltage on held-out flights from two excitation missions. The lower panels zoom into the highlighted load increases to show the effect of the load-onset term.}
\label{fig:identification}
\end{figure}

\begin{table}[b]
\centering\footnotesize
\caption{Physical model and training randomization}
\setlength{\tabcolsep}{3pt}
\begin{tabular}{@{}lll@{}}
\toprule
Quantity & Nominal & Randomization\\
\midrule
Mass $m$ [g] & $38$ & $\pm5\%$\\
$\operatorname{diag}\mathbf J$ [$10^{-5}$ kg m$^2$] & $3.3,3.6,5.9$ & $\pm20\%$\\
Rotor inertia $J_r$ [kg m$^2$] & $5\!\times\!10^{-8}$ & Fixed\\
Motor gain multiplier & $1$ & $\pm3/2\%$\\
Thrust multiplier & $1$ & $\pm5/3\%$\\
Motor lag $\tau_m$ [ms] & $50$ & $\pm15/5\%$\\
Torque multiplier & $1$ & $\pm10/5\%$\\
RP rate PID gains & $200,400,2.5$ & $\pm10\%$\\
Yaw rate PID gains & $120,16.7,0$ & $\pm10\%$\\
Attitude bias std. [deg] & $0.8,0.6,6$ & Clip $1.8,1.5,10$\\
Velocity bias mean [m/s] & $0,0,-.051$ & Std. $.02,.02,.005$\\
PID-history std. [deg/s] & $8,8,30$ & Clip $20,20,80$\\
\midrule
\multicolumn{3}{@{}l}{Battery parameters (fixed)}\\
$k_q$ [V/s]; $\rho$ & $7.0212\!\times\!10^{-4}$ & $1.25$\\
$(k_d,k_r)$ [V] & \multicolumn{2}{l}{$(0.03955847,0.10180777)$}\\
$(k_z,k_h)$ [V] & \multicolumn{2}{l}{$(0.04198630,0.02943918)$}\\
$(\tau_d,\tau_r,\tau_z,\tau_h)$ [s] & \multicolumn{2}{l}{$(0.10,0.02,6.0,0.30)$}\\
$(V_0,V_{\min},V_{\max})$ [V] & \multicolumn{2}{l}{$(4.20,2.80,4.25)$}\\
$\bar\Omega$ [rad/s]; update [s] & $2900$ & $0.01$\\
\bottomrule
\end{tabular}
\tabnote{Paired percentages give multiplicative perturbations shared by all motors, followed by those sampled per motor. RP denotes roll/pitch. Inertia comes from the identified rotor model, not mass alone. Mean attitude bias is $(0.3,0,0)$ deg. $\operatorname{diag}\mathbf K_{\rm aero}=(3.717,3.213,2.578)\!\times\!10^{-6}$, randomized by $\pm13.6/14.3/40\%$ along the three axes.}
\label{tab:plant}
\end{table}

Circle tracking uses a $1$\,m radius at $1.15$\,m height. The prescribed speeds are $3.36$ and $3.84$\,m/s, with $4.08$\,m/s as a stress test. These span precision tracking and limited thrust headroom: ideal collective demand rises from $0.57$ to $0.73$\,N before drag and attitude corrections (Fig.~\ref{fig:thrust_limits}). Preview offsets are $(0.05,0.10,0.20,0.30,0.45,0.65)$\,s, and reference radius and speed vary by $\pm2\%$ during training. Starts mix entry ramps and perturbed recovery states. Over the first $70\%$ of training, the ramp-start fraction decreases from $75\%$ to $50\%$ without changing reference difficulty. In~\eqref{eq:track_reward}, $(w_p,\lambda_p,k_1,k_2)=(0.1,0.75,5,50)$ and $(w_\psi,w_\omega,w_a)=(0.025,10^{-4},5\!\times\!10^{-4})$; tracking error above $1.25$\,m ends the episode.

\begin{figure*}[b]
\centering
\includegraphics[width=\textwidth]{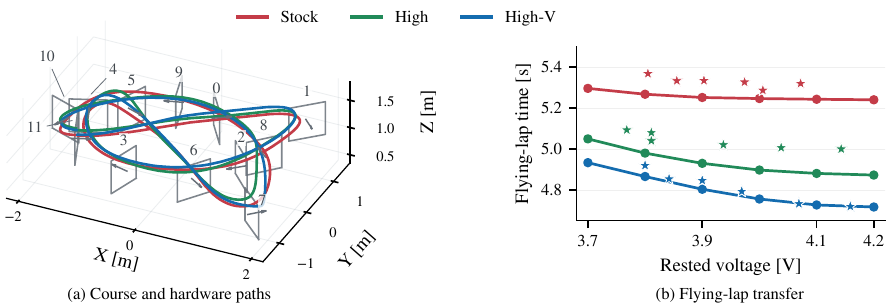}
\caption{Racing course and sim-to-real comparison. (a) Measured flying laps from flights starting near $4.0$\,V, with twelve $0.60$\,m virtual gates and arrows indicating the crossing direction. (b) Flying-lap times for the deployed policies in nominal simulation (lines) and hardware (stars).}
\label{fig:racing_summary}
\end{figure*}

The racing course has twelve upright $0.60$\,m virtual gates (Fig.~\ref{fig:racing_summary}(a)). Training uses a $0.35$\,m crossing window and gate-center jitter of $\pm0.05$\,m, with equal proportions of perturbed hover starts and recovery starts at $0.5$--$2.5$\,m/s tangent speed. In~\eqref{eq:race_reward}, $(w_d,w_g,w_r,w_c)=(2,0.5,2.5\!\times\!10^{-4},0.01)$ and $(c_0,c_{\max})=(0.20,0.75)$\,m. Both tasks use failure cost $10$ and arena bounds $[-2.4,2.4]\times[-2,2]\times[0.05,2]$\,m.

We evaluate final policies deterministically from official starts with nominal parameters, without domain randomization, over rested-equivalent reset voltages $3.70$--$4.20$\,V in $0.10$\,V steps.

For each method, we deploy the final policy with the best held-out simulation performance among those satisfying the safety criteria. Selection precedes hardware evaluation and minimizes mean RMSE or flying-lap time at $3.70/3.95/4.20$\,V, requiring failure and arena-exit rates below $1\%$ at each voltage. Primary selection uses three circle seeds and three racing seeds.

\begin{figure*}[t]
\centering
\includegraphics[width=\textwidth]{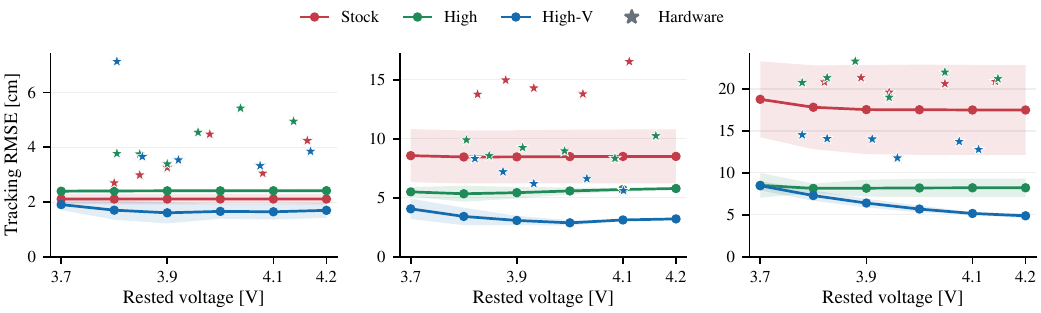}
\caption{Circle tracking at $3.36$, $3.84$, and $4.08$~m/s (left to right), with the last serving as a stress test. Hardware (stars) and nominal simulation (lines, shaded standard deviation) are both scored from $1.5$ to $10$\,s, after the entry ramp. All $50$ flights are shown. Preflight voltage uses the first four samples before motor loading.}
\label{fig:trajectory}
\end{figure*}

\subsection{Model Identification and Validation}\label{sec:identification}
We identify motor response and the base battery model from command steps and sustained loads with the vehicle rigidly mounted. Recorded duty, RPM, and terminal voltage give $\alpha=880.35$\,rad\,s$^{-1}$\,V$^{-\beta}$, $\beta=0.802$, and $\tau_m=50$~ms in~\eqref{eq:voltage_rotor}, without a current sensor.

Firmware compensation filters terminal voltage with $a_c=0.99$ at each $2$\,ms step (about $199$\,ms). Actor voltage instead uses a $54$\,ms filter at $100$\,Hz, optionally followed by a $5$ or $10$\,s filter at $50$\,Hz. These use~\eqref{eq:filter}, initialized from reset voltage.

Twenty electrical traces, eighteen with synchronized flight state, support the in-flight extension. Keeping the base model fixed, we fit the nonnegative load-onset coefficient by least squares, weighting recordings equally and sweeping its exponent and time constant (Table~\ref{tab:plant}). In leave-one-flight-out validation, this term reduces voltage RMSE from $50.2$ to $32.1$\,mV and high-load 95th-percentile absolute error from $161.9$ to $83.6$\,mV (Fig.~\ref{fig:identification}). Whole-flight holdouts avoid splitting correlated samples, although model structure was selected during this campaign. This validates voltage prediction, not the control benefit of each term.

\begin{figure*}[t]
\centering
\includegraphics[width=\textwidth]{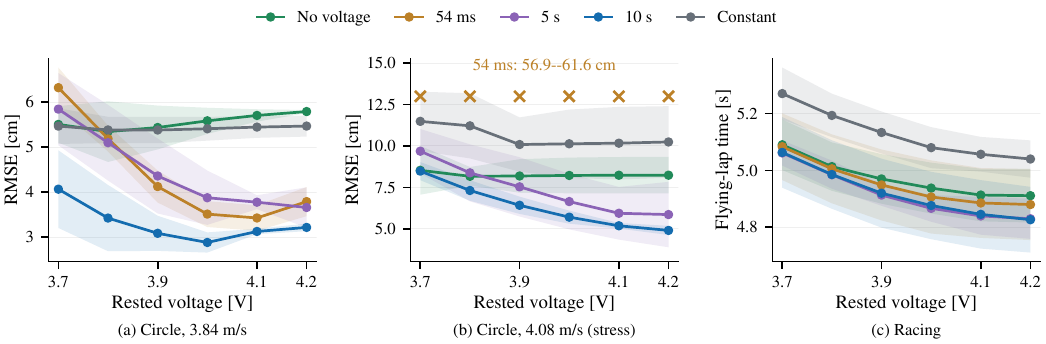}
\caption{Effect of voltage filtering at high authority. Legend entries give filter durations, with $10$\,s corresponding to \highv{}. Lines show means with one standard deviation shaded, using matched PPO and critic settings. Crosses mark $54$\,ms training collapse above the axis range. The constant input is $3.95$\,V.}
\label{fig:circle_voltage_features}
\end{figure*}

\begin{figure}[!b]
\centering
\includegraphics[width=\columnwidth]{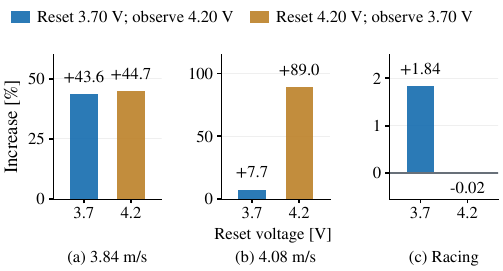}
\caption{\highv{} receives voltage recorded at the other battery condition, not a constant input. Bars show mean paired percentage increases in circle RMSE at (a) $3.84$ and (b) $4.08$~m/s (stress), and (c) racing lap time. Each panel uses its own linear scale.}
\label{fig:intervention}
\end{figure}

\subsection{Hardware Performance and Transfer}\label{sec:hardware}
We deploy the policies zero-shot in $50$ circle flights, with five or six per condition. Each lasts $20$\,s including a $1.5$\,s entry ramp, with rest periods between flights as voltage naturally changes.

\paragraph{Circle tracking}
Over the full $20$\,s flights, all three methods track at $3.36$\,m/s with approximately $4$\,cm mean error. At $3.84$\,m/s, RMSE falls from $14.26$\,cm for \stock{} to $8.53$\,cm for \high{} and $7.23$\,cm for \highv{}. Thus \highv{} achieves $15.3\%$ lower error than \high{}, which uses the same command range, and $49.3\%$ lower error than \stock{}.

At $4.08$\,m/s, \highv{} still tracks best, while \high{} falls behind \stock{}. Simulation underestimates error even over the matched $1.5$--$10$\,s window (Fig.~\ref{fig:trajectory}). A focused \high{} replay finds similar voltage and force-delivery ratios, but lower collective requests and more motor saturation in hardware. Estimator errors, controller history, and maneuver-dependent moments remain possible contributors to this closed-loop mismatch.

\paragraph{Gate racing}
All $27$ evaluated races complete the requested laps without failures or crashes. We report every completed race: six four-lap and three $20$-lap trials per method. The $4.8\times4.0$\,m motion-capture area constrains the course to short acceleration segments, tight turns, and height changes (Fig.~\ref{fig:racing_summary}(a)). Table~\ref{tab:racing_hardware} summarizes all $252$ laps using flight-level statistics.

\begin{table}[tb]
\centering\footnotesize
\caption{Hardware racing performance}
\setlength{\tabcolsep}{2.7pt}
\begin{tabular}{@{}lccc@{}}
\toprule
Metric & \stock{} & \high{} & \highv{} \\
\midrule
Four-lap race [s] & 21.30 $\pm$ 0.10 & 20.07 $\pm$ 0.16 & \textbf{19.26 $\boldsymbol{\pm}$ 0.27} \\
Flying lap, short [s] & 5.33 $\pm$ 0.03 & 5.04 $\pm$ 0.04 & \textbf{4.81 $\boldsymbol{\pm}$ 0.08} \\
20-lap race [s] & 106.22 $\pm$ 0.26 & 99.76 $\pm$ 0.84 & \textbf{95.24 $\boldsymbol{\pm}$ 0.44} \\
Flying lap, long [s] & 5.31 $\pm$ 0.01 & 4.99 $\pm$ 0.04 & \textbf{4.76 $\boldsymbol{\pm}$ 0.02} \\
Mean speed, long [m/s] & 3.74 & 3.89 & \textbf{4.10} \\
Peak speed, long [m/s] & 6.08 $\pm$ 0.11 & 6.74 $\pm$ 0.46 & \textbf{6.95 $\boldsymbol{\pm}$ 0.05} \\
\bottomrule
\end{tabular}

\tabnote{Flight-level means over six short and three long trials per method, with standard deviation (SD) where shown. Bold marks the best value in each row. Flying laps omit the standing start, whereas total race time includes it. Peak speed averages the maximum of each long flight.}
\label{tab:racing_hardware}
\end{table}

\begin{figure}[b]
\centering
\includegraphics[width=\columnwidth]{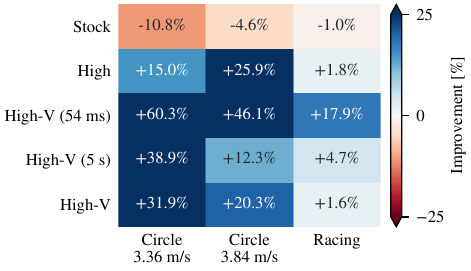}
\caption{Mean paired percentage reduction in error or lap time with a privileged rather than symmetric critic, where positive values indicate improvement. Parentheses identify alternative voltage filters, and colors are capped at $\pm25\%$.}
\label{fig:critic}
\end{figure}

\begin{table}[t]
\centering\footnotesize
\caption{Critic prediction accuracy}
\label{tab:critic_fit}
\renewcommand{\arraystretch}{1.12}
\setlength{\tabcolsep}{3.5pt}
\begin{tabular*}{\columnwidth}{@{\extracolsep{\fill}}llcc@{}}
\toprule
Task & Actor & Symmetric & Privileged \\
\midrule
\multirow{3}{*}{\shortstack[l]{Circle\\3.36\,m/s}} & \stock{} & 0.93 $\pm$ 0.01 & \textbf{0.99 $\boldsymbol{\pm}$ 0.01} \\
 & \high{} & 0.94 $\pm$ 0.01 & \textbf{0.97 $\boldsymbol{\pm}$ 0.02} \\
 & \highv{} & \textbf{0.98 $\boldsymbol{\pm}$ 0.01} & \textbf{0.98 $\boldsymbol{\pm}$ 0.01} \\
\midrule
\multirow{3}{*}{\shortstack[l]{Circle\\3.84\,m/s}} & \stock{} & 0.96 $\pm$ 0.00 & \textbf{0.97 $\boldsymbol{\pm}$ 0.01} \\
 & \high{} & 0.96 $\pm$ 0.00 & \textbf{0.98 $\boldsymbol{\pm}$ 0.00} \\
 & \highv{} & \textbf{0.99 $\boldsymbol{\pm}$ 0.00} & \textbf{0.99 $\boldsymbol{\pm}$ 0.00} \\
\midrule
\multirow{3}{*}{\shortstack[l]{Racing}} & \stock{} & 0.19 $\pm$ 0.02 & \textbf{0.81 $\boldsymbol{\pm}$ 0.10} \\
 & \high{} & 0.18 $\pm$ 0.00 & \textbf{0.86 $\boldsymbol{\pm}$ 0.03} \\
 & \highv{} & 0.71 $\pm$ 0.05 & \textbf{0.81 $\boldsymbol{\pm}$ 0.03} \\
\bottomrule
\end{tabular*}

\tabnote{Mean $\pm$ SD across runs, after averaging each run's final 20 logged explained-variance values. Higher is better; bold marks the largest mean at the displayed precision, including ties.}
\end{table}

Over $20$ laps, \high{} and \highv{} reduce \stock{}'s time by $6.1\%$ and $10.3\%$, respectively; \highv{} is $4.5\%$ faster than \high{}. Its mean per-flight peak speed reaches $6.95$\,m/s, versus $6.08$\,m/s for \stock{}. For the deployed checkpoints, nominal simulation predicts mean flying-lap time within approximately $0.10$\,s of hardware (Fig.~\ref{fig:racing_summary}(b)).

All three methods brake at tight reversals (Fig.~\ref{fig:racing_summary}(a)). \highv{} follows a similar, slightly longer route than \high{}, gaining time through higher mean speed rather than a shorter path. A larger area might allow greater use of its additional authority.

Replaying recorded commands through the actuator model gives mean errors of $407$--$614$ PWM counts and $2.3$--$3.6\%$ in rotor speed. \stock{} has the lowest electrical replay error, while \highv{} has the closest agreement between simulated and measured lap times.

\subsection{Which Voltage Input Helps?}\label{sec:information}
We compare voltage filters at fixed high authority to choose an input that works across both tasks (Fig.~\ref{fig:circle_voltage_features}). Short and slow filters improve circle tracking at $3.84$\,m/s, whereas a constant $3.95$\,V input stays close to \high{}. Adding an input dimension alone therefore does not reproduce the benefit. At the $4.08$\,m/s stress speed, however, training with the $54$\,ms filter is less reliable, with off-scale error marked in the figure.

We retain the $10$\,s filter in \highv{} to provide slowly varying battery-condition information while onboard compensation handles faster voltage changes. This may reduce sensitivity to short load fluctuations. Neither the $54$\,ms nor $5$\,s alternative meets our replacement criteria: at least $3\%$ improvement on both the $3.84$\,m/s circle and racing, no more than $3\%$ regression at $3.36$\,m/s, and failure below $1\%$.

We test voltage use with paired simulations from identical initial conditions. One receives the correct voltage and the other a recording from the opposite battery condition ($3.70$ or $4.20$\,V). The policy and physical battery stay unchanged, and both runs evolve freely.

With the battery initialized at $4.20$\,V, supplying the $3.70$\,V recording substantially worsens circle tracking (Fig.~\ref{fig:intervention}), while the reverse swap has a more variable effect. In racing, incorrect voltage increases lap time by $1.84\%$ at $3.70$\,V but has approximately zero mean effect at $4.20$\,V. The learned response to voltage therefore depends on both the task and battery condition.

\subsection{Does Battery Information Help the Critic?}\label{sec:critic}
Battery information substantially improves the critic's return prediction in racing (Table~\ref{tab:critic_fit}). Circle critics already explain most return variation, especially at $3.84$\,m/s, leaving less room for improvement in prediction accuracy.

In Fig.~\ref{fig:critic}, asymmetric training improves mean performance for every high-authority variant, while \stock{} regresses. Battery information may stabilize policy updates by explaining return differences from changing thrust availability. For \stock{}, conservative commands and compensation may make this information less useful, while added critic complexity could hinder optimization.

\section{CONCLUSIONS}\label{sec:conclusions}
We presented a battery-aware RL approach that combines higher thrust-command authority with a filtered voltage input while retaining onboard compensation and rate control. Its training model connects battery discharge and load-induced voltage drops to the forces available to the policy.

On the Crazyflie Brushless, the approach nearly halves tracking RMSE on the harder operational circle and reduces $20$-lap race time by $10.3\%$ relative to stock-authority policies. It also improves on voltage-blind policies with the same increased command range. Together with the voltage-input tests, these results show that explicit voltage feedback can complement the compensation already provided by the flight controller.

Future work will combine voltage feedback with motion-history adaptation and extend the model to current and power prediction, as in battery-constrained NMPC~\cite{gupta2026bc}. This would allow policies to balance aggressive flight with energy consumption.

\section*{ACKNOWLEDGMENT}
OpenAI ChatGPT/Codex and Anthropic Claude helped draft and revise the abstract,
Sections I--V and figure captions, and write analysis and plotting code.
The authors reviewed, edited and verified all such content and take full
responsibility.

\end{document}